\documentclass{article}

\usepackage{times}
\usepackage{graphicx} 

\usepackage{subcaption}

\usepackage{natbib}

\usepackage[algo2e,ruled]{algorithm2e}

\usepackage{hyperref}

\usepackage[accepted]{icml2016}

\usepackage{eqnarray,amsmath}
\usepackage{amssymb}

\usepackage{amsthm}

\usepackage{pifont}

\usepackage{paralist}

\DeclareMathOperator*{\argmax}{\arg\!\max}

\newcommand*\rot{\rotatebox{45}}

\theoremstyle{definition}

\icmltitlerunning{Fast Optimization of Wildfire Suppression Policies with SMAC}

\begin{document}

\twocolumn[
\icmltitle{Fast Optimization of Wildfire Suppression Policies with SMAC}

\icmlauthor{Sean McGregor}{arXiv@seanbmcgregor.com}
\icmladdress{School of Electrical Engineering and Computer Science,
            Oregon State University}
\icmlauthor{Rachel Houtman}{rachel.houtman@oregonstate.edu}
\icmladdress{College of Forestry,
            Oregon State University}
\icmlauthor{Claire Montgomery}{claire.montgomery@oregonstate.edu}
\icmladdress{College of Forestry,
            Oregon State University}
\icmlauthor{Ronald Metoyer}{rmetoyer@nd.edu}
\icmladdress{College of Engineering,
            University of Notre Dame}
\icmlauthor{Thomas G. Dietterich}{tgd@oregonstate.edu}
\icmladdress{School of Electrical Engineering and Computer Science,
            Oregon State University}

\icmlkeywords{Markov Decision Processes, Monte Carlo Methods, Surrogate Modeling, Visualization, SMAC, Bayesian Policy Search}

\vskip 0.3in
]


\begin{abstract}
    Managers of US National Forests must decide what policy to apply for dealing with lightning-caused wildfires. Conflicts among stakeholders (e.g., timber companies, home owners, and wildlife biologists) have often led to spirited political debates and even violent eco-terrorism. One way to transform these conflicts into multi-stakeholder negotiations is to provide a high-fidelity simulation environment in which stakeholders can explore the space of alternative policies and understand the tradeoffs therein. Such an environment needs to support fast optimization of MDP policies so that users can adjust reward functions and analyze the resulting optimal policies. This paper assesses the suitability of SMAC---a black-box empirical function optimization algorithm---for rapid optimization of MDP policies. The paper describes five reward function components and four stakeholder constituencies. It then introduces a parameterized class of policies that can be easily understood by the stakeholders. SMAC is applied to find the optimal policy in this class for the reward functions of each of the stakeholder constituencies. The results confirm that SMAC is able to rapidly find good policies that make sense from the domain perspective.  Because the full-fidelity forest fire simulator is far too expensive to support interactive optimization, SMAC is applied to a surrogate model constructed from a modest number of runs of the full-fidelity simulator. To check the quality of the SMAC-optimized policies, the policies are evaluated on the full-fidelity simulator. The results confirm that the surrogate values estimates are valid. This is the first successful optimization of wildfire management policies using a full-fidelity simulation. The same methodology should be applicable to other contentious natural resource management problems where high-fidelity simulation is extremely expensive.
\end{abstract}

\section{Introduction}



When lightning ignites a fire in the US National Forest system, the forest manager must decide whether to suppress that fire or allow it to burn itself out. This decision has immediate costs in terms of fire fighting expenses and smoke pollution and long-term benefits, including increased timber harvest revenue and reduced severity of future fires. Different stakeholders place different values on these various outcomes, and this leads to contentious and difficult policy debates.  In the US Pacific Northwest, a period in the 1990s is referred to as the ``Timber Wars'' because of the troubling and occasionally violent conflicts that arose between stakeholder groups over forest management policies during that period. This is typical of many ecosystem management problems---the complexity of ecosystem dynamics and the broad array interested parties makes it difficult to identify politically-feasible policies. 

One way that computer science can help is to provide a high-fidelity simulation environment in which stakeholders can explore the policy space, experiment with different reward functions, compute the resulting optimal policies, and visualize the behavior of the ecosystem when it is managed according to those policies. This process can elicit missing aspects of the reward function, and it can help the stakeholders reach a policy consensus that is informed by the best available ecosystem models.  To create such a simulation environment, we need a tool that meets the following requirements:
\begin{enumerate}[i.]
    \item \textbf{Modifiability}: users should be able to modify the reward function to represent the interests of various stakeholders.
    \item \textbf{Automatic Optimization}: users should be able to optimize policies for the updated reward functions without the involvement of a machine learning researcher.
    \item \textbf{Visualization}: users should be able to explore
    the behavior of the system when it is controlled by the optimized policies.
    \item \textbf{Interactivity}: all these tasks should be
    performed at interactive speeds.
\end{enumerate}

Previous work by McGregor et al.~\yrcite{McGregor2015a} presented an interactive visualization system, MDPvis, that supports requirements i and iii. However, MDPvis does not support the optimization capability needed for requirement ii. Furthermore, the full-fidelity wildfire simulator is very slow, so even if we had an optimization algorithm for this noisy, high-dimensional problem, the optimization could not meet the interactive speeds needed for requirement iv.

This paper and a companion paper \cite{McGregor2017a} (simultaneously published on arXiv) address requirements ii and iv. The companion paper shows how to create a surrogate model that is able to accurately mimic the expensive full-fidelity simulator while running orders of magnitude faster. This addresses requirement iv.  The current paper studies whether the SMAC method for black-box function optimization \cite{Hutter2010} can use this surrogate model to rapidly find near-optimal policies. 

SMAC is similar to Bayesian methods for black box optimization. However, unlike those methods, SMAC does not employ a Gaussian process to model the black box function. Instead, it fits a random forest ensemble \cite{Breiman2001}. This has three important benefits. First, it does not impose any smoothness assumption on the black box function. We will see below that wildfire policies are naturally expressed in the form of decision trees, so they are highly non-smooth. Second, SMAC does not require the design of a Gaussian process kernel. This makes it more suitable for application by end users such as our policy stakeholders. Third, the CPU time required for SMAC scales as $O(n \log n)$ where $n$ is the number of evaluations of the black box function, whereas standard GP methods scale as $O(n^3)$ because they must compute and invert the kernel matrix. 

This paper makes two contributions.  First, it shows that SMAC can rapidly find high-scoring policies for a range of different reward functions that incorporate both short-term and long-term rewards. Second, it confirms that this is possible even though SMAC is using an approximate surrogate for the high-fidelity simulator.  Taken together, these contributions mark the first successful optimization a wildfire suppression policy for a
full 100-year planning horizon.

The paper is structured as follows. First, we introduce our notation and provide an overview of direct policy search and SMAC. This is followed by a description of the fire management problem including a review of the different components of the reward function and the relative weight that different stakeholder constituencies place on these components. We then describe the parameterized policy representation for wildfire management policies. The results of applying SMAC to optimize these policies are shown next. Finally, the surrogate estimates of the values of these policies are checked by running them on the full-fidelity simulator. The results confirm the accuracy of the surrogate estimates. 

\section{Direct Policy Search Methods} \label{sec:related}

We work with the standard finite horizon discounted MDP
\cite{Bellman1957,Puterman1994}, denoted by the tuple
$M=\langle{S,A,P,R,P_0,h,\gamma\rangle}$. $S$ is a finite set of states of
the world; $A$ is a finite set of possible actions that can be taken
in each state; $P: S\times A\times S \mapsto [0,1]$ is the conditional
probability of entering state $s'$ when action $a$ is executed in
state $s$; $R(s,a)$ is the finite reward received after performing
action $a$ in state $s$; $P_0$ is the distribution over starting
states; $h$ is the horizon; and $\gamma$ is the discount factor.

Let $\Pi$ be a class of deterministic policies with an associated parameter space $\Theta$. Each parameter vector $\theta \in \Theta$ defines a policy $\pi_{\theta}: S \mapsto A$ that specifies what action to take in each state.  Let $\tau= \langle s_0, s_1, \ldots, s_h\rangle$ be a trajectory generated by drawing a state $s_0 \sim P_0(s_0)$ according to the starting state distribution and then following policy $\pi_{\theta}$ for $h$ steps. Let $\rho = \langle r_0, \ldots, r_{h-1}\rangle$ be the corresponding sequence of rewards. Both $\tau$ and $\rho$ are random variables because they reflect the stochasticity of the starting state and the probability transition function. Let $V_{\theta}$ define the expected cumulative discounted return of applying policy $\pi_{\theta}$ starting in a state $s_0 \sim P_0(s_0)$ and executing it for $h$ steps: 
\[
V_{\theta} = \mathbb{E}_\rho[r_0+\gamma r_1 + \gamma^2 r_2 + \cdots + \gamma^{h-1} r_{h-1}]
\]
The goal of direct policy search is to find $\theta^*$ that maximizes the value of the corresponding policy:
\[
\theta^* = \argmax_{\theta \in \Theta} V_{\theta}.
\]
Two prominent forms of policy search are policy gradient methods and sequential model-based optimization. Policy gradient methods \cite{Williams1992,Sutton2000,Deisenroth2011,Schulman2015} estimate the gradient of $V_\theta$ with respect to $\theta$ and then take steps in parameter space to ascend the gradient. This is often challenging because the estimate is based on Monte Carlo samples of $\tau$ and because gradient search only guarantees to find a local optimum. 

Sequential model-based optimization methods \cite{Kushner1964,Mockus1994,Zilinskas1992,Hutter2010,Srinivas2010,Wilson2014,Wang2016} construct a model of $V_\theta$ called a Policy Value Model and denoted $PVM(\theta)$. The PVM estimates both the value of $V_\theta$ and a measure of the uncertainty of that value. The most popular form of PVM is the Gaussian Process, which models $V_\theta$ as the GP mean and the uncertainty as the GP variance. The basic operation of sequential model-based optimization methods is to select a new point $\theta$ at which to improve the PVM, observe the value of $V_\theta$ at that point (e.g., by simulating a trajectory $\tau$ using $\pi_\theta$), and then update the PVM to reflect this new information. In Bayesian Optimization, the PVM is initialized with a prior distribution over possible policy value functions and then updated after each observation by applying Bayes rule. The new points $\theta$ are selected by invoking an {\it acquisition function}. 

SMAC \cite{Hutter2010} is a sequential model-based optimization method in which the PVM is a random forest of regression trees. The estimated value of $V_\theta$ is obtained by ``dropping'' $\theta$ through each of the regression trees until it reaches a leaf in each tree and then computing the mean and the variance of the training data points stored in all of those leaves. In each iteration, SMAC evaluates $V_\theta$ at 10 different values of $\theta$, adds the observed values to its database $R$ of $(\theta, V_\theta)$ pairs, and then rebuilds the random forest. 

SMAC chooses 5 of the 10 $\theta$ values with the goal of finding points that have high ``generalized expected improvement''. The (ordinary) expected improvement at point $\theta$ is the expected increase in the maximum value of the PVM that will be observed when we measure $V_\theta$ under the assumption that $V_\theta$ has a normal distribution whose mean is $\mu_\theta$ (the current PVM estimate of the mean at $\theta$) and whose variance is $\sigma^2_\theta$ (the PVM estimate of the variance at $\theta$). The expected improvement at $\theta$ can be computed as
\begin{equation} \label{eq:acquisition}
 EI(\theta) := \mathbb{E}\big[I(\theta)\big] = \sigma_{\theta}\big[z\cdotp\Phi(z)+\phi(z)\big],
\end{equation}
where $z:=\frac{\mu_{\theta}-f_{max}}{\sigma_{\theta}}$,
$f_{max}$ is the largest known value of the current PVM, 
$\Phi$ denotes the cumulative distribution function of the standard
normal distribution, and $\phi$ denotes the probability density function of the standard normal distribution \cite{Jones1998}.

The generalized expected improvement (GEI) is obtained by computing the expected value of $I(\theta)$ raised to the $g$-th power. In SMAC, $g$ is set to 2. \citet{Hutter2010} show that this can be computed as 
\begin{equation} \label{eq:acquisition2}
 GEI(\theta) = \mathbb{E}\big[I^2(\theta)\big] = \sigma_{\theta}^2\big[(z^2+1)\cdotp\Phi(z)+z\cdotp\phi(z)\big].
\end{equation}

Ideally, SMAC would find the value of $\theta$ that maximizes $GEI(\theta)$ and then evaluate $V_\theta$ at that point. However, this would require a search in the high-dimensional space of $\Theta$ and it would also tend to focus on a small region of $\Theta$. Instead, SMAC employs the following heuristic strategy to find 10 candidate values of $\theta$. First, it performs a local search in the neighborhood of the 10 best known values of $\theta$ in the PVM. This provides 10 candidate values. Next, it randomly generates another 10,000 candidate $\theta$ vectors from $\Theta$ and evaluates the GEI of each of them.  Finally, it chooses the 5 best points from these 10,010 candidates and 5 points sampled at random from the 10,000 random candidates, and evaluates $V_\theta$ at each of these 10 points. This procedure mixes ``exploration'' (the 5 random points) with ``exploitation'' (the 5 points with maximum GEI), and it has been found empirically to work well. 

\citet{Hutter2010} prove that the SMAC PVM is a consistent estimator of $V$ and that given an unbounded number of evaluations of $V$, it finds the optimal value $\theta^*$. 

\section{The Wildfire Management Domain} \label{sec:vis_wildfire}

\begin{figure}
    \centering
        \includegraphics[width=0.95\columnwidth]{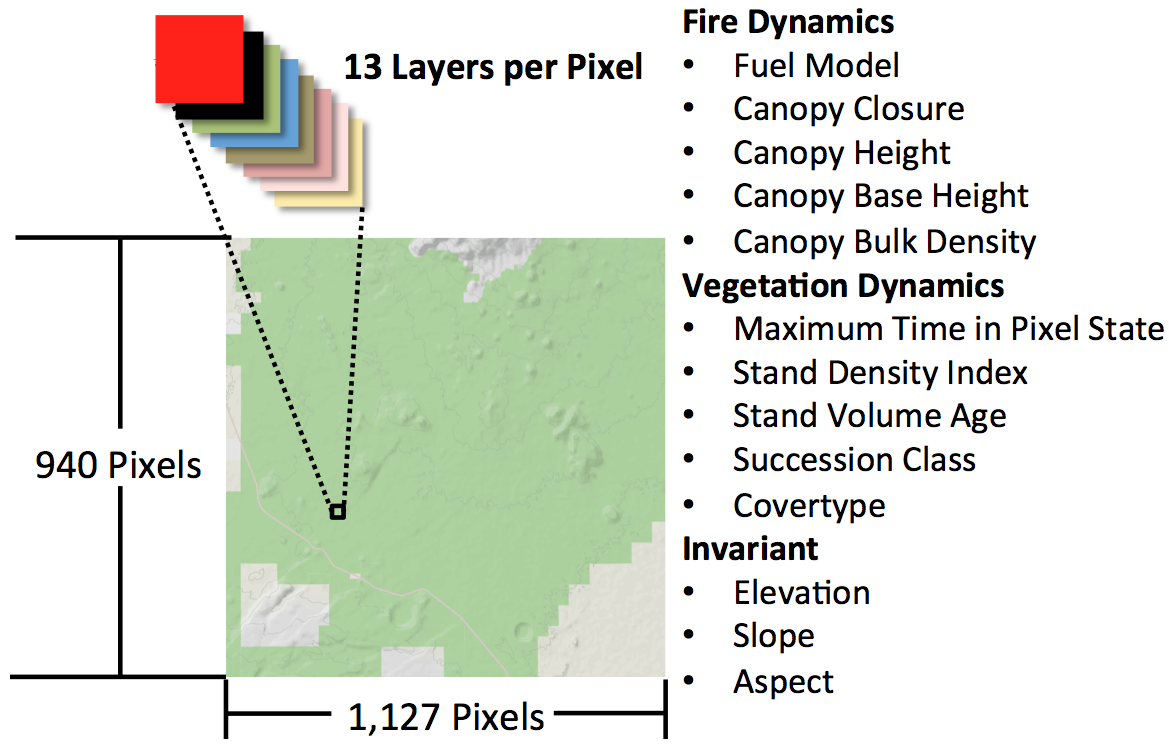}
        \caption{The landscape totals approximately one million
          pixels, each of which has 13 state variables that influence
          the spread of wildfire on the landscape.
          (Map is copyright of OpenStreetMap contributors)}
        \label{fig:pixel-layers}
\end{figure}

To evaluate our methods, we have selected a portion of the Deschutes National Forest in Eastern Oregon. This area was selected because it is being managed with the goal of restoring the landscape to the condition it was believed to be in prior to the arrival of Europeans. Figure \ref{fig:pixel-layers} shows a map of the study site. It is comprised of approximately one million pixels, each described by 13 state variables.

We employ the high-fidelity simulator described in \citet{Houtman2013}, which combines a simple model of the spatial distribution of lightning strikes (based on historical data) with the state-of-the-art Farsite fire spread simulator \cite{Finney1998}, a fire duration model \cite{finney2009modeling}, and the high-resolution FVS forest growth simulator \cite{Dixon2002}. Weather is simulated by resampling from the historical weather time series observed at a nearby weather station.

The MDP advances in a sequence of decision points. Each decision point corresponds to a lightning-caused ignition, and the MDP policy must decide between two possible actions: \textit{Suppress} (fight the fire) and \textit{Let Burn} (do nothing). Hence, $|A|=2$. Based on the chosen action and the (simulated) weather, the intensity, spread, and duration of the fire is determined by the simulator. In order to capture the long-term impacts of our policy decisions, we employ a planning horizon of $h=100$ years.

Unfortunately, the simulator is very expensive. Simulating a 100-year trajectory of fires can take up to 7 hours of clock time. This is obviously too slow for interactive use.  We therefore have adopted a surrogate modeling approach in which, during the optimization, we replace the full-fidelity model with a much more efficient approximation. Our surrogate model is based on the Model Free Monte Carlo (MFMC) method of \citet{Fonteneau2013}.  MFMC generates trajectories for a new policy by combining short segments of trajectories previously computed by the full simulator for other policies. It relies on a similarity metric to match segments to each other. Because this metric does not work well in a high-dimensional state-and-action space, the companion paper describes an extension of MFMC called MFMCi (MFMC with independencies) that exploits certain conditional independencies to reduce the dimensionality of the similarity metric calculations \cite{McGregor2017a}. 

Wildfires produce a variety of immediate and long term losses and
benefits. For market-based rewards, such as suppression costs and
timber revenues, there is a single defensible reward function. Since
wildfire decisions also affect many outcomes for which there is no
financial market, such as air, water, ecology, and recreation, there
are potentially many different composite reward functions.  A benefit
of the random forest method is the ability to change the reward
function and re-optimize without making any assumptions about the character of the
response surface $V$.  In our experimental evaluation, we optimize and
validate policies for four different reward functions as an
approximation of different stakeholder interests. Table
\ref{table:rewards} details each of these reward function
constituencies.

{
  \begin{table}[ht]
    \centering
    \footnotesize
    \begin{tabular}{p{1.85cm}p{0.4cm}p{0.4cm}p{0.4cm}p{0.4cm}p{0.4cm}}
      Constituency & \rot{Suppression Costs} & \rot{Timber Revenues} & \rot{Ecology Target} & \rot{Air Quality} & \rot{Recreation Target} \\ [0.5ex]
      \hline
      Composite & \checkmark & \checkmark & \checkmark & \checkmark & \checkmark\\
      Politics & - & \checkmark & \checkmark & \checkmark & \checkmark\\
      Home Owners & - & - & - & \checkmark & \checkmark\\
      Timber & \checkmark & \checkmark & - & - & - \\[1ex]
      \hline
    \end{tabular}
    \caption{ Components of each reward function. The
      ``politics'' constituency approximates a decision maker that
      is not responsible for funding firefighting operations. The
      ``home owner'' constituency only cares about air quality and
      recreation.  The ``timber'' companies only care about how much
      timber they harvest, and how much money they spend protecting
      that timber. The ``composite'' reward function takes an unweighted
      sum of all the costs and revenues produced for the constituencies. 
      Additional reward functions can be specified by users interactively
      within MDPvis.}
    \label{table:rewards}
\end{table}}

The reward functions are compositions of five different reward components. The \textit{Suppression} component gives the expenses incurred for suppressing a fire.
Fire suppression expenses increase with fire size and the number of
days the fire burns.
Without fire suppression effort, the fire suppression costs are zero,
but the fire generally takes longer to self-extinguish.
\textit{Timber} harvest values are determined by the number of board feet
harvested from the landscape. A harvest scheduler included in the simulator
determines the board feet based on existing forest practice regulations.
Generally we would expect timber harvest to increase with
suppression efforts, but complex interactions between the harvest scheduler and
tree properties (size, age, species) often results in high timber harvests following
a fire.
\textit{Ecological} value is a function of the squared
deviation from an officially-specified target distribution of vegetation on the landscape known as the ``restoration target.''
Since our landscape begins in a state that has a recent history of fire suppression efforts, there is much more vegetation than the target. A good way to reach the target is to allow wildfires to burn, but increased timber harvest can also contribute to this goal.
\textit{Air Quality} is a function of the number of days a wildfire
burns. When fires burn, the smoke results in a large number of home-owner complaints. We encode this as a negative reward for each smoky day.
Finally, the \textit{recreation} component penalizes the squared deviation from a second vegetation target distribution---namely, one preferred by people hiking and camping. This distribution consists of old, low-density ponderosa pine trees. Frequent, low-intensity fires produce this kind of distribution, because they burn out the undergrowth while leaving the fire-resilient ponderosa pine unharmed. If we optimize for any single reward component, the optimal policy will tend to be one of the trivial policies ``suppress-all'' or ``letburn-all''.  When multiple reward components are included, the optimal policy still tends to either suppress or let burn most fires by default, but it tries to identify exceptional fires where the default should be overridden. See \citet{Houtman2013} for a discussion of this phenomenon.

\begin{figure}
    \centering
        \includegraphics[width=0.95\columnwidth]{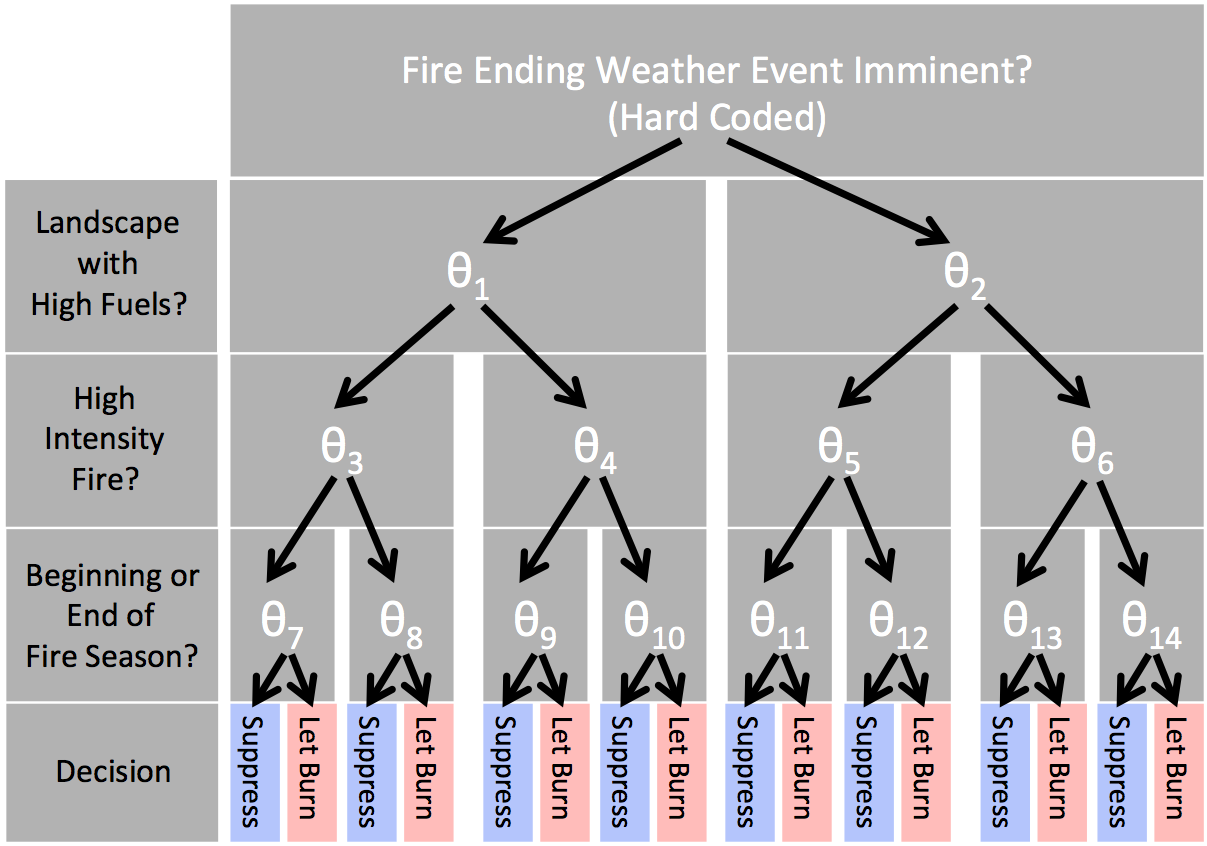}
        \caption{ The layers of the decision tree used to select
          wildfires for suppression.  The top layer splits on whether
          the fire will likely be extinguished by a storm in the next
          8 days regardless of the suppression decision.  The next
          layers then have 14 parameters for the number of pixels that
          are in high fuel (parameters 1 and 2, $\big[0,1000000\big]$), the intensity of the weather
          conditions at the time of the fire (3 through 6, $\big[0,100\big]$), and a
          threshold that determines whether the fire is close to
          either the start or end of the fire season (7 through 14, $\big[0,90\big]$).
        }
        \label{fig:decision_tree}
\end{figure}

A natural policy class in this domain takes the form of a binary decision tree as shown in Figure \ref{fig:decision_tree}. At each level of the tree, the variable to split on is fixed in this policy class. With the exception of the very first split at the root, which has a hard-coded threshold, the splitting thresholds $\theta_1,\ldots,\theta_{14}$ are all adjusted by the policy optimization process. Moving from the top layer of the tree to the bottom, the tree splits first on whether the fire will be extinguished within the next 8 days by a ``fire-ending weather event'' (i.e., substantial rain or snowfall). The threshold value of 8 is fixed (based on discussions with land managers and on the predictive scope of weather forecasts).  The next layer splits on the state of fuel accumulation on the landscape. The fuel level is compared either to $\theta_1$ (left branch, no weather event predicted within 8 days) or $\theta_2$ (right branch; weather predicted within 8 days). When the fuel level is larger than the corresponding threshold,
the right branch of the tree is taken. The next layer splits on the intensity of the fire at the time of the ignition. In this fire model, the fire intensity is quantified in terms of the Energy Release Component (ERC), which is a common composite measure of dryness in fuels. Finally, the last layer of the tree asks whether the current date is close to the start or end of the fire season. Our study region in Eastern Oregon is prone to late spring and early fall rains, which means fires near the boundary of the fire season are less likely burn very long. We note that
this policy function is difficult for gradient-based policy search, because it is not differentiable and exhibits complex responses to parameter changes.

\section{Experiments}

To train the MFMCi surrogate model, we sampled 360 policies from a policy class that
suppresses wildfires based on the ERC at the time of the ignition and the day of the ignition (see supporting materials for details). Every query to the MFMCi surrogate generates 30 trajectories, and the mean cumulative discounted reward is returned as the observed value of $V_\theta$. 

We apply SMAC with its default configuration. When SMAC grows a random regression tree for its PVM, there is a parameter that specifies the fraction of the parameter dimensions (i.e., $\theta_1, \ldots, \theta_{14}$) that should be considered when splitting a regression tree node. We set this parameter to $5/6$. A second parameter determines when to stop splitting, namely: a random forest node can only be split if it contains at least 10 examples. Finally, the size of the random forest is set to 10 trees.

\begin{figure*}
    \centering
    \begin{subfigure}[t]{0.47\textwidth}
        \includegraphics[width=0.9\textwidth]{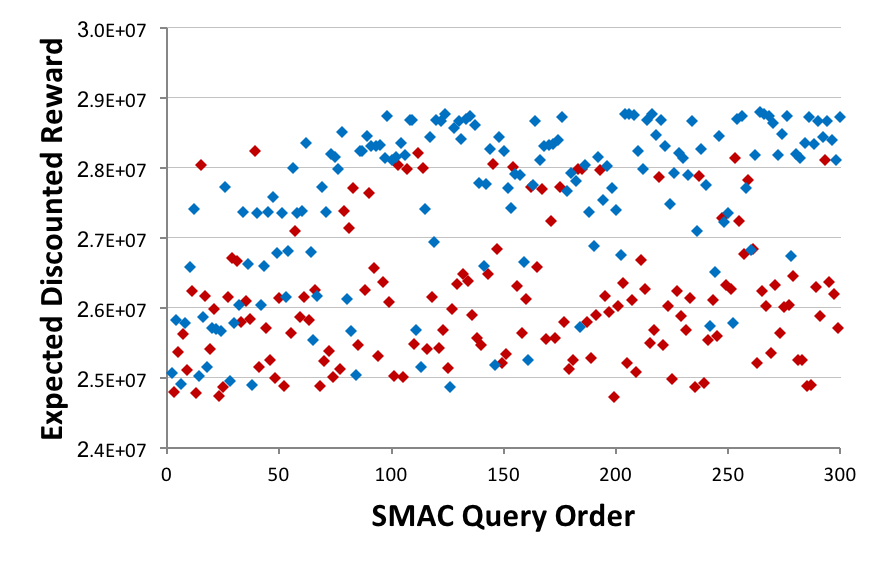}
        \caption{
        Composite reward function.
        }
        \label{fig:composite_exploration}
    \end{subfigure}
    \hfill
    \begin{subfigure}[t]{0.47\textwidth}
        \includegraphics[width=0.9\textwidth]{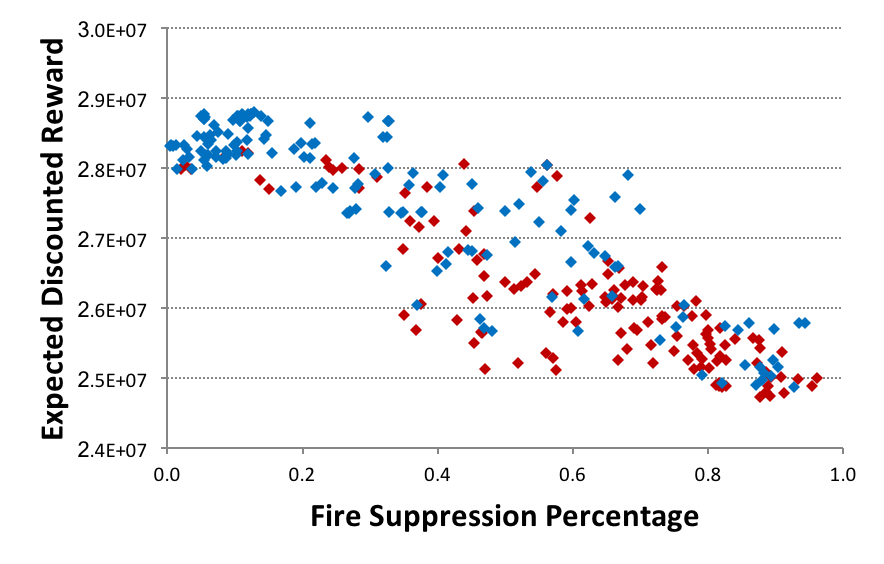}
        \caption{
        Composite reward function.
        }
        \label{fig:composite_suppression_count}
    \end{subfigure}

    \begin{subfigure}[t]{0.47\textwidth}
        \includegraphics[width=0.9\textwidth]{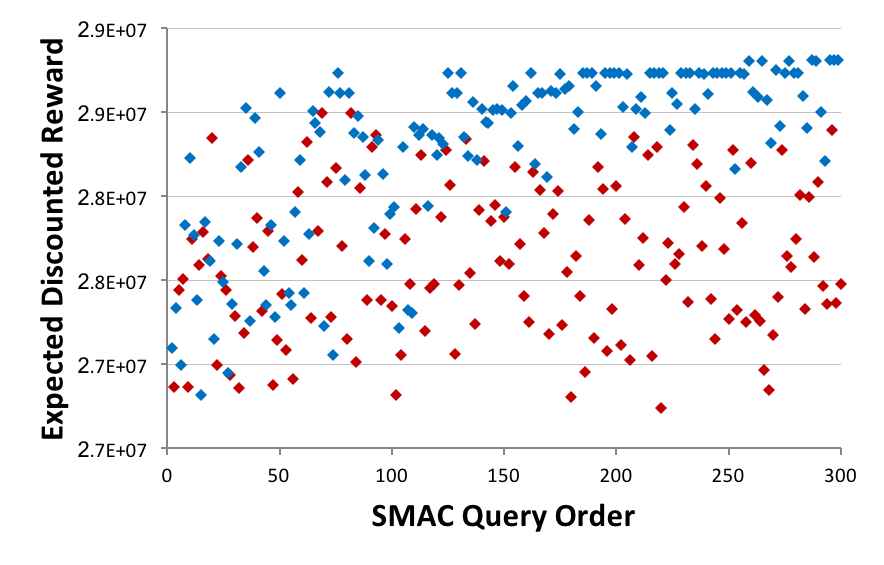}
        \caption{
        Politics reward function.
        }
        \label{fig:politics_exploration}
    \end{subfigure}
    \hfill
    \begin{subfigure}[t]{0.47\textwidth}
        \includegraphics[width=0.9\textwidth]{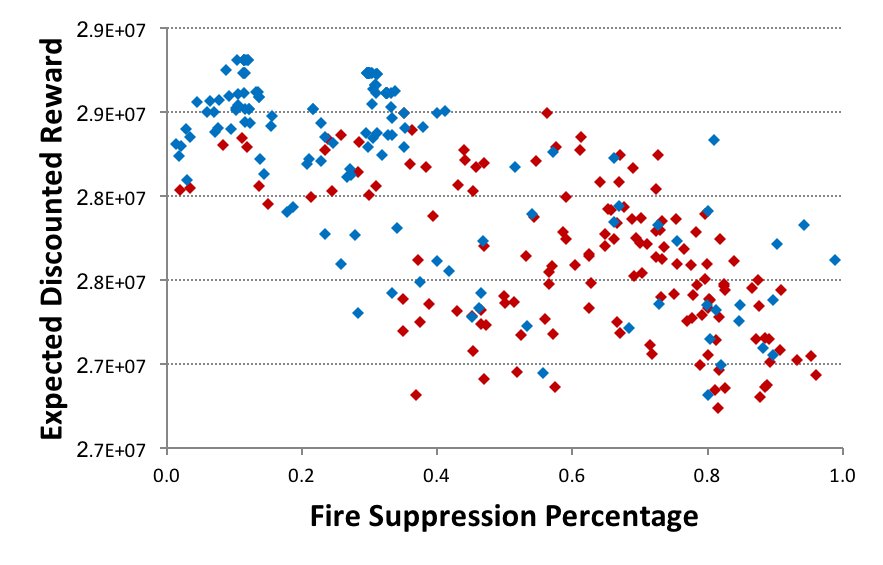}
        \caption{
        Politics reward function.
        }
        \label{fig:politics_suppression_count}
    \end{subfigure}

    \begin{subfigure}[t]{0.47\textwidth}
        \includegraphics[width=0.9\textwidth]{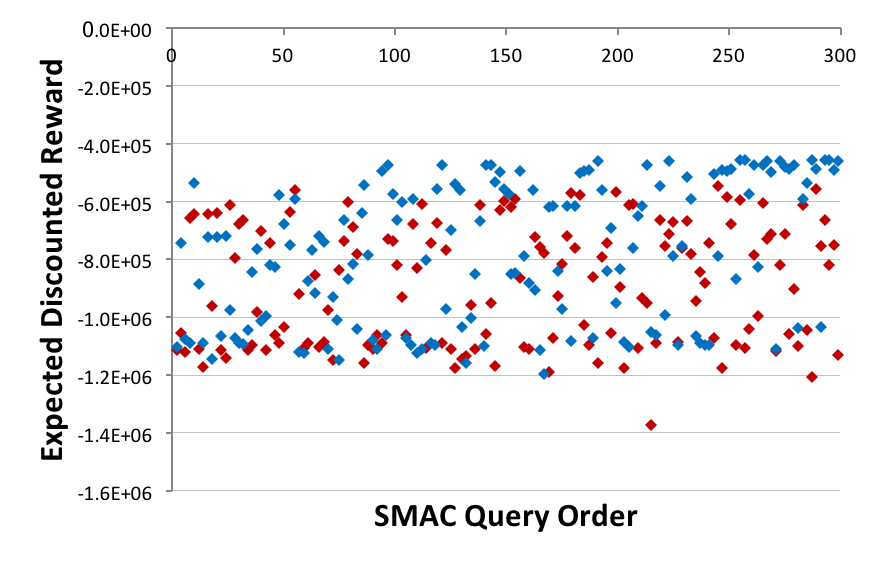}
        \caption{
        Home owner reward function.
        }
        \label{fig:home_owners_exploration}
    \end{subfigure}
    \hfill
    \begin{subfigure}[t]{0.47\textwidth}
        \includegraphics[width=0.9\textwidth]{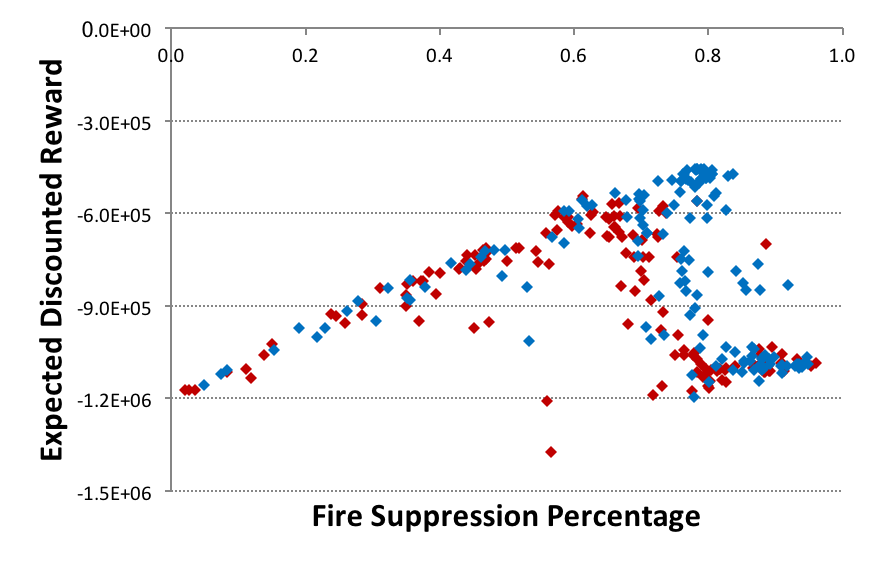}
        \caption{
        Home owners reward function.
        }
        \label{fig:home_owners_suppression_count}
    \end{subfigure}

    \begin{subfigure}[t]{0.47\textwidth}
        \includegraphics[width=0.9\textwidth]{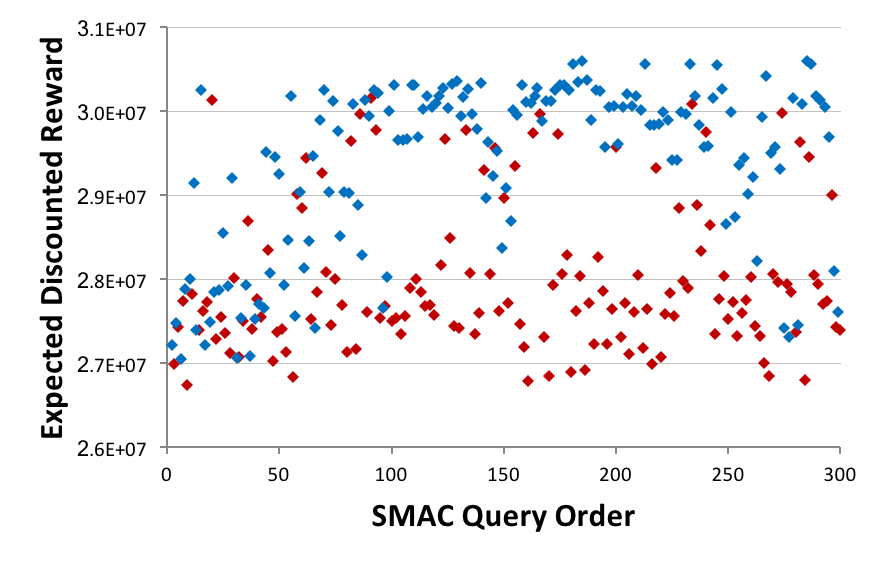}
        \caption{
        Timber reward function.
        }
        \label{fig:timber_exploration}
    \end{subfigure}
    \hfill
    \begin{subfigure}[t]{0.47\textwidth}
        \includegraphics[width=0.9\textwidth]{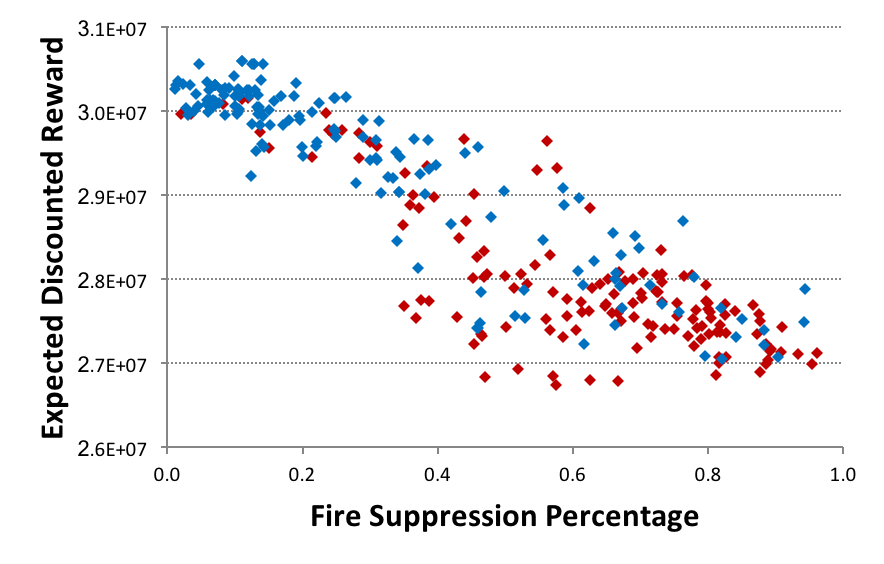}
        \caption{
        Timber reward function.
        }
        \label{fig:timber_suppression_count}
    \end{subfigure}
    \caption{ Discounted reward achieved for each reward function
      averaged over 30 trajectories.  The blue diamonds are samples
      selected according to the EI heuristic, and the red diamonds are
      randomly sampled points.  }
    \label{fig:parameter_exploration}
\end{figure*}

Figure \ref{fig:parameter_exploration} shows the results of applying SMAC to find optimal policies for the four reward function constituencies. The left column of plots show the order in which SMAC explores the policy parameter space. The vertical axis is the estimated cumulative discounted reward, and the point that is highest denotes the final policy output by SMAC. Blue points are policy parameter vectors chosen by the GEI acquisition function whereas red points are parameter vectors chosen by SMAC's random sampling process. Notice that in all cases, SMAC rapidly finds a fairly good policy. The right column of plots gives us some sense of how the different policies behave. Each plot sorts the evaluated policy parameter vectors according to the percentage of fires that each policy suppresses.  In \ref{fig:parameter_exploration}(b), we see that the highest-scoring policies allow almost all fires to burn, whereas in  \ref{fig:parameter_exploration}(f), the highest-scoring policies suppress about 80\% of the fires. 

Let us examine these policies in more detail. The optimal policies for the \textit{politics} and \textit{timber} reward constituencies allow most wildfires to burn, but for different reasons. For the \textit{politics} constituency, it is the Ecological reward that encourages this choice, whereas for the \textit{timber} constituency, it is the increased harvest levels that result. The \textit{composite} reward function produces a very similar optimal policy, presumably because it contains both the Ecological and Harvest reward components. These results indicate that timber company and political interests largely coincide.

The most interesting case is the \textit{home owner} constituency reward function, which seeks to minimize smoke (which suggests suppressing all fires) and maximize recreation value (which suggests allowing fires to burn the understory occasionally). We can see in \ref{fig:parameter_exploration}(f) that the best policy found by SMAC allows 20\% of fires to burn and suppresses the remaining 80\%.

\begin{figure*}
    \centering
    \begin{subfigure}[t]{0.49\textwidth}
        \includegraphics[width=1.0\textwidth]{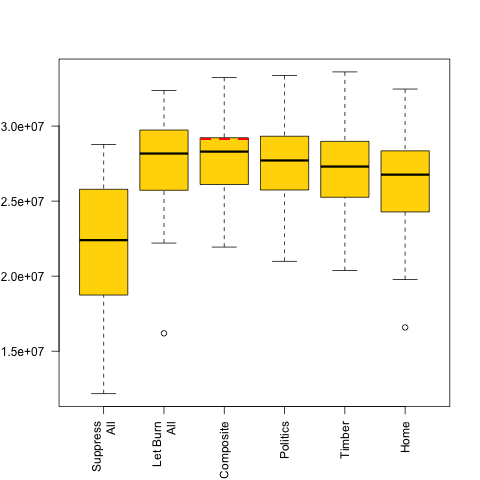}
        \caption{
        Policies for composite reward function.
        }
        \label{fig:composite_violin}
    \end{subfigure}
    \hfill
    \begin{subfigure}[t]{0.49\textwidth}
        \includegraphics[width=1.0\textwidth]{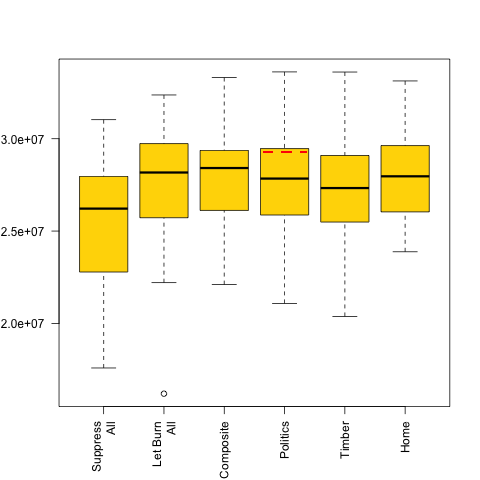}
        \caption{
        Policies for politics reward function.
        }
        \label{fig:politics_violin}
    \end{subfigure}

    \begin{subfigure}[t]{0.49\textwidth}
        \includegraphics[width=1.0\textwidth]{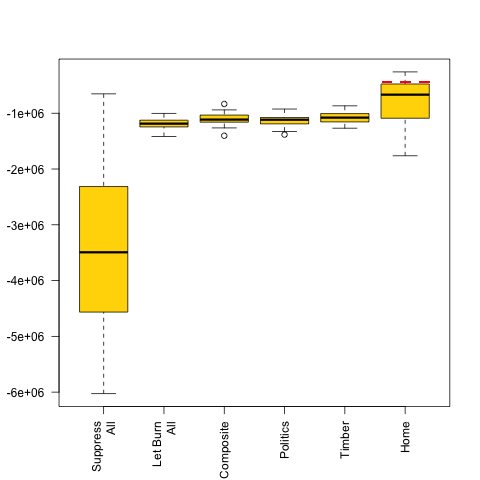}
        \caption{
        Policies for home owner reward function.
        }
        \label{fig:home_violin}
    \end{subfigure}
    \hfill
    \begin{subfigure}[t]{0.49\textwidth}
        \includegraphics[width=1.0\textwidth]{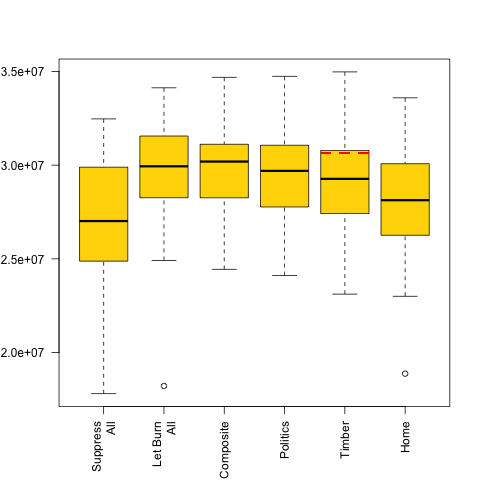}
        \caption{
        Policies for timber reward function.
        }
        \label{fig:timber_violin}
    \end{subfigure}
    \caption{
    Each set of box charts show the performance of various policies for a constituency. The individual box
    plots show the expected discounted reward for each of the policies optimized for
    a constituency, as well as the hard-coded policies of suppress-all and let-burn-all.
    The red dashed lines indicate the expected discounted reward estimated by MFMCi.}
    \label{fig:symphony}
\end{figure*}
These results agree with our intuition for each reward constituency and provide evidence that SMAC is succeeding in optimizing these policies.  However, the expected discounted rewards in Figure \ref{fig:parameter_exploration} are estimates obtained from the modified MFMC surrogate model. To check that these estimates are valid, we invoked each optimal policy on the full simulator at least 50 times and measured the cumulative discounted reward under each of the reward functions. We also evaluated the Suppress All and Let Burn All policies for comparison. The results are shown in Figure \ref{fig:symphony}.  

Each panel of boxplots depicts the range of cumulative returns for each policy on one reward function. For the policy that SMAC constructed, we also plot a dashed red line showing the MFMC estimate of the return. In all cases, the estimates are systematically biased high. This is to be expected, because any optimization against a noisy function will tend to fit the noise and, hence, over-estimate the true value of the function. Nonetheless, the MFMC estimates all fall within the inter-quartile range of the full simulator estimates.  This confirms that the MFMC estimates are giving an accurate picture of the true rewards.

Note that because of the stochasticity of this domain, using only 50 trajectories from the full simulator is in general not sufficient to determine which policy is optimal for each reward function. The only clear case is for the \textit{home owner} reward function where the SMAC-optimized policy is clearly superior to all of the other policies.

\section{Discussion}

Previous work on high-fidelity wildfire management \cite{Houtman2013} has focused only on {\it policy evaluation}, in which the full simulator was applied to evaluate a handful of alternative policies. This paper reports the first successful {\it policy optimization} for wildfire suppression at scale. This paper demonstrates that SMAC applied to our MFMC-based surrogate model is able to find high-quality policies for four different reward functions and do so at interactive speeds.  This combination of optimization efficiency and robust ease of use has the potential to provide a basis for interactive decision support that can help diverse groups of stakeholders explore the policy ramifications of different reward functions and perhaps reach consensus on wildfire management policies. 

\section*{Acknowledgment}
  This material is based upon work supported by the National Science Foundation under Grant No. 1331932.

\bibliographystyle{icml2016}
\bibliography{references}

\end{document}